\documentclass[conference, sort]{IEEEtran}
\IEEEoverridecommandlockouts

\usepackage{amsmath,amssymb,amsfonts}
\usepackage{algorithmic}
\usepackage{graphicx}
\usepackage{placeins}
\usepackage{textcomp}
\usepackage{xcolor}
\usepackage{fancyvrb}
\usepackage[hidelinks]{hyperref}

\def\BibTeX{{\rm B\kern-.05em{\sc i\kern-.025em b}\kern-.08em
    T\kern-.1667em\lower.7ex\hbox{E}\kern-.125emX}}

\usepackage{caption}

\usepackage[numbers]{natbib}

\begin{document}

\title{Logic Mill - A Knowledge Navigation System
\thanks{For helpful comments, we thank Matt Marx, as well as participants at the 2022 Munich Summer Institute and the 2022 Summer School on Data and Algorithms for Science, Technology \& Innovation Studies at KU Leuven. 
We also thank the team at the computing and IT competence center of GWDG for their continuous support.
Dietmar Harhoff acknowledges support from Deutsche Forschungsgemeinschaft (CRC 190).}}

\author
{\IEEEauthorblockN{
Sebastian Erhardt, Mainak Ghosh, Erik Buunk, Michael E. Rose, Dietmar Harhoff}
\IEEEauthorblockA{\textit{Max Planck Institute for Innovation and Competition} \\Munich, Germany\\
Email: \{sebastian.erhardt, mainak.ghosh, erik.buunk, michael.rose, dietmar.harhoff\}@ip.mpg.de
}
}

\maketitle

\begin{abstract}
\textit{Logic Mill} is a scalable and openly accessible software system that identifies semantically similar documents within either one domain-specific corpus or multi-domain corpora. 
It uses advanced Natural Language Processing (NLP) techniques to generate numerical representations of documents.
Currently it leverages a large pre-trained language model to generate these document representations.
The system focuses on scientific publications and patent documents and contains more than 200 million documents.
It is easily accessible via a simple Application Programming Interface (API) or via a web interface.
Moreover, it is continuously being updated and can be extended to text corpora from other domains. 
We see this system as a general-purpose tool for future research applications in the social sciences and other domains.
\end{abstract}


\section{Introduction}

There is a growing need for tools that allow researchers to identify related documents within the same, but also across different domains. 
With the ever-growing volume of scientific publications and patents, scholars find it burdensome to manage relevant documents and search important prior contributions efficiently.
Finding relevant documents plays a significant role in building coherent scientific arguments, but is also important in assessing the use of scientific research outside academia \cite{marx_reliance_2020, poege_science_2019}. 
Patent examination is another field in which finding related documents and identifying prior art is essential.
In ex post analyses, researchers often rely on citation data to identify relations between documents. While citations are helpful in tracing citation networks and in uncovering important patterns in the production and diffusion of knowledge within the same corpus, they are typically limited when searching for relations across different corpora. 
Even within the same corpus, citations can be selective or even  systematically biased (see \cite{jaffe_patent_2017, rubin_systematic_2021}). Finally, references may not exist for texts that are in the process of being created, so that authors are faced with the challenge of identifying relevant references in the first place. Therefore, tools are needed that allow for processing and analyzing the textual contents of high-volume text corpora and of establishing measures of relatedness (similarity) between them.

We refer to the system described here as a knowledge navigation system, since it allows for tracing knowledge elements within and across text corpora (e.g., the corpus of all patents or of all scientific publications). Previous attempts include \cite{woltmann_tracing_2018} for scientific publications and \cite{kelly_measuring_2021} for patent documents, both of make use of bag-of-words approaches. 
While such systems are sometimes referred to as recommender systems \cite{beel_research-paper_2016}, recommendation of related documents is only one possible application of knowledge navigation (see below for a non-exhaustive list of use cases). 
However, these systems often use proprietary algorithms, usually focus on one domain corpus, are not openly accessible, or are not continuously being updated.
A knowledge navigation system of the kind envisioned here should be capable of efficiently retrieving, storing, and processing hundreds of millions of documents. 
Moreover, it requires capabilities for fast detection of particularly similar documents within and across corpora.

An important problem that needs to be addressed is how to implement the concept of document similarity. 
This requires representing documents in a numerical form that computers can process, with the goal of generating similar representations for similar documents at large scale. 
In the field of natural language processing (NLP) this process is known as document encoding.

There are various methods for representing documents numerically.
Traditional NLP approaches like \textit{TF-IDF} \cite{sparck_jones_statistical_1972} are used widely in the literature.
However, these traditional methods are not scalable, since an extension of the vocabulary requires a re-computation of all representations of the corpus.
In addition, these approaches do not capture the semantics of the documents; that is the meaning of words, or the interpretation of sentences in context, is lost.

Therefore we propose \textit{Logic Mill}, a software system aiming to satisfy the shortcomings of existing systems and approaches. 
The system creates document representations using modern NLP techniques and contains large document sets with pre-calculated encodings.
It is easy to use and allows users to access and compare texts from different text corpora.
Furthermore, it is scalable and built on non-proprietary algorithms. We regularly update the datasets based on their release schedule. 
Our objective is to provide a fast system of high accuracy that is openly accessible.

In the release version, the system encodes text documents using the \textit{SPECTER} document encoder \cite{cohan_specter_2020} which leverages the bi-directional transformer architecture (BERT) \cite{devlin_bert_2019}.
This model, in combination with the database containing numerical representations of documents from different corpora (analogous to a vector search database), is the backbone for the \textit{Logic Mill} system.



At present, we provide the numerical representations for the scientific articles of Semantic Scholar and for the patent publications issued by the United States Patent and Trademark Office (USPTO), the European Patent Office (EPO), and the World Intellectual Property Organization (WIPO). To be precise, we use the titles and abstracts of these, since BERT-based models are restricted to vectors of 1,024 tokens.

An important feature is that users can also feed their own text data to Logic Mill for encoding and for obtaining similarity measures, both within their data and between own data and standard text data of the system. 
They can thus link their own curated documents to patents and scientific publications according to textual similarity.

\textit{Logic Mill} can be used in a number of different research applications, such as:
\begin{itemize}
\item Explore literature:
Search for research papers, and find the best matches based on textual similarity to a paper in the database or to own text documents.

\item Prior art search in patent examination:
Look for previously granted patents or (not yet granted) patent applications, but are similar to the focal one.


\item Link patents to related scientific publications:
Search for patents that the scientific publication might be based on or have a strong similarity to.

\item Recommend citations and readings for new documents:
Find documents that are very similar to a focal one and may be useful as a reference or reading.

\item Assess the novelty of patents and publications:
Check if a patent or publication is new or not by comparing it to prior texts. 
Documents that have few highly similar documents may be new or even unique.

\item Trace concepts across domains and over time:
Identify documents across domains (e.g. publications and patents) that are highly similar and possibly related.  

\end{itemize}
\section{Document Encoding}

Document encoding in natural language processing (NLP) is a process for representing textual data in a numerical form.
There are various approaches to encode documents.

\paragraph{Bag of words}
Simple and fast procedures construct a vector whose binary elements indicate the presence of a word in a document ("binary term encoding"), the number of occurrences of a word in a document ("count matrix") or the weighted number of occurrences of a word in a document ("Term frequency-inverse document frequency (TF-IDF)").
These approaches have the drawback of not capturing the meaning or context of the words in the document -- hence the common term bag-of-words. 
Furthermore, they are not scalable since the whole model must be retrained if a new word is added to the vocabulary and the length of each vector equals with the vocabulary size. 
From a computational perspective, they are inefficient since they generate sparse matrices where most elements equal 0.

\paragraph{Word embeddings} 
Word embeddings are dense, fixed-size, and continuous-valued vector representations of words that capture the meaning of the words in the document. 
These word representations are learned via training over large corpora of textual data using methods such as \textit{Word2Vec} \cite{mikolov_efficient_2013}, \textit{GloVe} \cite{pennington_glove_2014}, or \textit{FastText} \cite{bojanowski_enriching_2017}. 
The advent of these word embedding methods was a leap towards memory-efficient dense numerical representation for words and documents from the bag of words models' sparse representation.
A baseline approach to representing a document is to average or sum the learned word embeddings of the words in a document.

\paragraph{Sentence/Paragraph Embeddings} 
Sentence embeddings can be understood as an extension of the basic idea of word embeddings.
Word embeddings are static representation of words and do not change even in the presence of multiple contexts in the document collection. However, the same word can have different meanings in different contexts. 
For example, the word \textit{"bank"} can be a financial institute or can relate to a river bank. 
Hence, the neural network architecture such as Recurrent Neural Networks (\textit{RNNs}) and Long Short-Term Memory (\textit{LSTM}) \cite{hochreiter_long_1997} are used for representing the word to its true context dynamically with a notion that words appearing either before or after the focal word reveal the context around the focal word. 
\textit{RNN} and \textit{LSTM} also aid the vector representation of the context of the sentences or paragraph \cite{melamud_context2vec_2016}.

However, \textit{RNN} and \textit{LSTM} architectures cannot appropriately capture the meaning of the words at the beginning of a very long sentence or paragraph in its numerical representation due to their sequential structure. 
The rise of deep neural networks and recent advancements in the field of NLP introduced the transformer architecture \cite{vaswani_attention_2017}. Transformers looks at each word of a sentence together, unlike RNN and LSTM, and learns the degree to which the words reflect the context of that sentence using the so-called attention mechanism. 
The transformer architecture can translate the meaning of each word in a sentence through its network into the numerical representation of the sentence, also called sentence embedding.
As stated before, a baseline approach to representing a document is to average or sum the learned sentence embeddings of that document.
 
\paragraph{BERT Language Model} 
Bidirectional Encoder Representations from Transformer, \textit{BERT} \cite{devlin_bert_2019} is another recent development that uses transformer architecture. 
It exhibits all the traits of transformer architecture, meaning it learns the sense of the words of an input sentence, the context of that sentence, and the semantic relation between the words and the context.
\textit{BERT} being a "bidirectional" model, considers the context of a word from both sides (left and right) at the same time in a sentence, which makes this model effectively process long contiguous text sequences, such as entire paragraph, not limited to short phrase.
The \textit{BERT} architecture is designed with a limitation of 512 input tokens. 
This means that before the text is fed into the model, it is tokenized and truncated if necessary. 
Any model built on \textit{BERT} must take this into account. 
Based on our own data, a word usually consists of $\approx 1.2$ tokens.

\hyphenation{SciBERT}
\paragraph{SciBERT} 
\textit{SciBERT} is a \textit{BERT} language model for tasks involving scientific publications \cite{beltagy_scibert_2019}. 
It was trained on a large corpus of 1.14M scientific publications from computer science and the broad biomedical field. This model also outputs numerical representation like \textit{BERT}.

Fine-tuning a general purpose \textit{BERT} model on scientific papers produces more accurate result in this domain \cite{beltagy_scibert_2019, aslanyan_patents_2022}, because it uses a domain-specific vocabulary. This also extends to similar domains, in this case patents, where \textit{SciBERT} model outperforms the original \textit{BERT} for tasks such as IPC classification and similar patent finding \cite{althammer_linguistically_2021}. \textit{SciBERT} is thus particularly well-suited for tasks such as information extraction, document classification, and text representation in the scientific domain.

\paragraph{SPECTER}
\textit{SPECTER} is an extension of \textit{SciBERT} to encode scientific publications also with the help of inter-document relatedness \cite{cohan_specter_2020}. In the scientific literature, citations signal relatedness, but this information is not used by \textit{SciBERT}. \textit{SPECTER} transfers the learned relatedness signal to the representation of a scientific article. 
During the application of this model, it generates similar embeddings for related scientific documents without knowing citation information.

We use \textit{SPECTER} model as our workhorse document encoder model.\footnote{There is one other \textit{BERT}-based language model specifically trained on patents, namely \textit{PatentBERT} \cite{lee_patent_2020}. Its task is to classify patents, and therefore not suitable for our purpose.} 
It generates embeddings for scientific publications and patents using all available information based on a domain-specific pre-learned vocabulary, even without citation information during the encoding process. 
Because BERT has a limitation of 512 tokens, only the title and abstract will be used as input. 
Based on our data, most of the tile + abstract is below the 512 token limitation.

\section{System specifics}

\begin{figure*}[ht!]
  \caption{Logic Mill Architecture Overview\label{fig:lm_architecture}}
  \includegraphics[width=18cm]{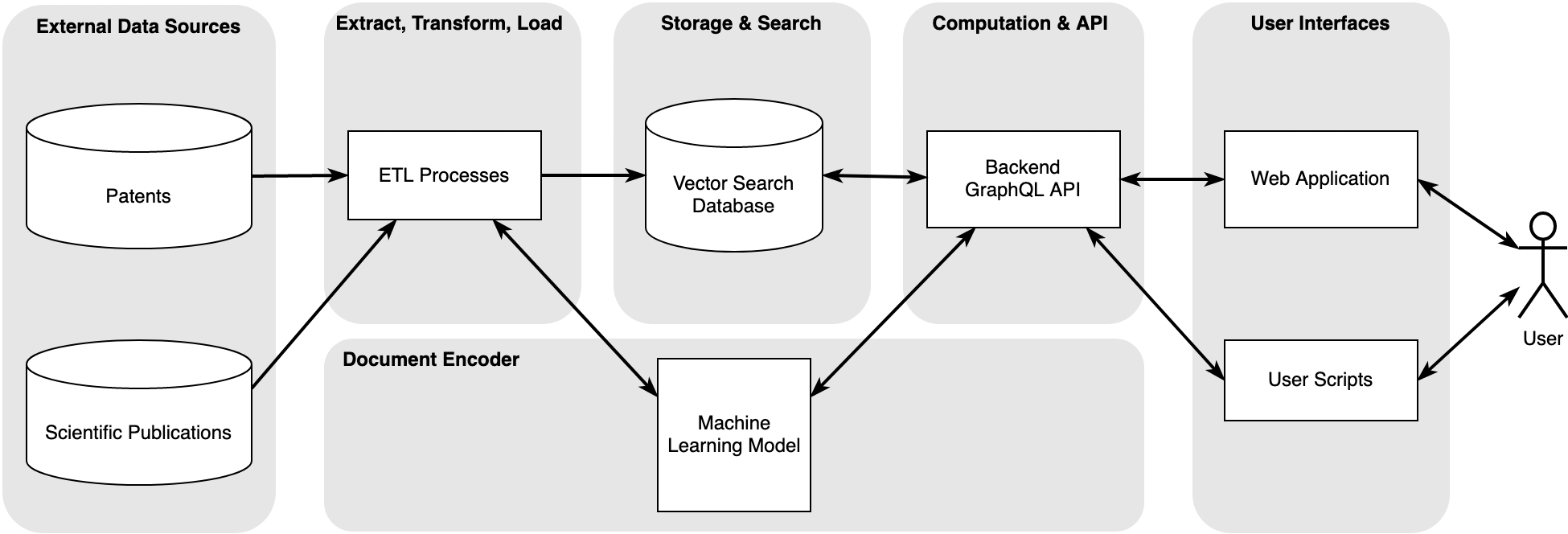}
\end{figure*}

\textit{Logic Mill} system is designed using a Microservice Architecture. 
This software design approach breaks down a large, monolithic application into smaller, independent components that can be developed, deployed, and maintained separately. 
Each microservice is a self-contained unit that performs a specific function and communicates with other microservices through well-defined interfaces, typically using APIs.

Figure \ref{fig:lm_architecture} shows the multiple services of our system.
From a high-level perspective, there are 6 distinct parts:
\begin{itemize}
    \item[a)] {External Data Sources}: where patent documents and scientific publications are obtained from
    \item[b)] \textit{Document Encoder}: transforms the text of the document into a numerical representation using a machine learning model
    \item[c)] \textit{Extract, Transform, Load (ETL)}: processes move documents from the external sources, process and store them
    \item[d)] \textit{Vector Search Database}: stores the computed numerical representation along with metadata and the text
    \item[e)] \textit{Backend with Web API}: propagate the user requests to the database
    \item[f)] \textit{User-Interface}: provides the users with a web application where they can interact with the API via our website or using their own scripts, for example in \textit{Python} or \textit{R}.
\end{itemize}

\paragraph{External Data Sources}
The system can be connected to various external data sources.
The release version of the system retrieves patent documents and scientific publications from public and access-restricted sources.

Scientific publications are obtained from the Semantic Scholar dataset, including abstracts and additional metadata such as publication date, journal name, or Digital Object Identifier (DOI) \cite{ammar_construction_2018}. 
The bulk dataset of Semantic Scholar is published as JSON files and contains more than 200 million documents. 
We automatically retrieve the latest version every month.

We obtain patent documents from multiple sources, namely the European Patent Office (EPO), the United States Patent and Trademark Office (USPTO), and the World Intellectual Property Organization (WIPO). 

The EPO offers various data sets and data feeds.\footnote{See \url{https://www.epo.org/searching-for-patents/data.html}}
We use DocDB and the API of the European Publication Server (EPS)\footnote{See \url{https://data.epo.org/publication-server/}.} to obtain full text and metadata.
DocDB includes bibliographic data from over 100 countries worldwide, and for some patent authorities, the data goes back as far as the 1830s.
The EPS API gives online access to all the European patent documents published by the EPO, which \textit{Logic Mill} retrieves as XML feeds.
Currently the system contains more than 7 million full-text patent documents from the EPO EPS and 145 million metadata records from the EPO DocDB.

We also retrieve XML files from the USPTO containing the full text and all relevant metadata.
Currently, there are more than 10 million patent documents, which are obtained through the USPTO Bulk Api. \footnote{See \url{https://developer.uspto.gov/api-catalog}.}
Currently the system contains more than 10 million full-text patent documents issued by the USPTO.

WIPO likewise provides XML files for the full text of international patents with metadata. 
Presently, there are around 3.8 million patent documents in the system, which were obtained via the WIPO's file server.

\textit{Logic Mill} retrieves continuously the latest patent documents from the respective sources once per week and feeds them into the system automatically.

\paragraph{Extract, Transform, Load}
Through various Extract, Transform, and Load (ETL) processes, the system obtains the raw documents from external data sources and processes them.
In the first step, the textual content and the metadata are extracted and stored in a global data structure. The structure is independent of the document type (patent or scientific publication).
In the second step, the document encoder encodes the text parts of the document and generates the numerical representation.
Finally, the search database stores the numerical representation of a document along with the metadata and the full text.

\paragraph{Document Encoder}
The document encoder is machine-learning model that transforms text documents into a numerical representation. We use \textit{SPECTER} \cite{cohan_specter_2020} to encode documents.
The output of this model is a dense vector with 768 dimensions.

Since the encoding of all documents requires significant computing power, the encoding was conducted on desktop workstations with Nvidia graphics processing units (GPU) and in high-performance cloud computing facilities.
To allow for real-time inference, a CPU container was deployed in the cloud and is connected to the system and accessible for end-users via the API.

\paragraph{Storage \& Search}
For the search and the storage of documents with their numerical representation ElasticSearch is used.
This database is capable of full-text search and can be used in a distributed context, which is essential for scalability reasons. 

Finally, ElasticSearch allows storing dense vectors that nearest-neighbor search algorithms can use.
Exact Nearest Neighbor searchers are guaranteed to find a solution, but are inefficient and not scalable.\footnote{They would require to pre-compute all distance metrics between the query vector and every vector in the database. In our setting this amounts to 220M+ pairs.}
Therefore we use approximate Nearest Neighbor searches (ANN), which trade-off precision for lower computational and resource burden. 
ElasticSearch uses the Hierarchical Navigable Small World graphs (HNSW) \cite{malkov_efficient_2018} algorithm (as of version 8.0).\footnote{Compared to a wide spectrum of alternative ANN and according to various distance measures, HNSW performs consistently well \cite{aumuller_ann-benchmarks_2020}; see also \url{http://ann-benchmarks.com}.} 
HNSW organizes vectors as a graph based on their similarity to each other. 
Together, this setup finds the most similar documents with very high probability for any query document within milliseconds.\footnote{We intend to provide further information on the time-accuracy trade-off in future research.}

In our current setting the cluster consists of $12$ nodes with the 8 vCPU cores, 128 GB of RAM, and 1 TB of SSD storage each. 
This setting is needed to allow for fast and efficient computation, because the RAM required by the ElasticSearch database can be distributed over multiple nodes. 
The database cluster as well as all other components are running on the GWDG OpenStack Cloud IT infrastructure\footnote{Gesellschaft für wissenschaftliche Datenverarbeiting mbH Göttingen, see \url{https://www.gwdg.de/}.} in Göttingen, Germany.

\paragraph{Computation \& API}
The back-end extends the software stack for more functionality and to handle the user interactions.
It is written in the language \textit{Go} and provides a plug-and-play Application Programming Interface (API) for end-users using \textit{GraphQL}.
This query language for APIs is a strongly typed interface that provides complete and understandable API documentation.
Furthermore, it allows the users to retrieve the data precisely that they have asked for.

The back-end also connects the document encoder with the client-facing run-time environment.
Doing that ensures that end-users can send text, which can then be encoded to the numerical representation, and finally, this representation can be used to query similar documents from the database.

It can also be used to calculate distances between texts provided by the user using their numerical representation and distance metrics like the Cosine Similarity (Eq. \eqref{cos}), the Manhattan (L1) Distance (Eq. \eqref{l1}) or the Euclidean (L2) Distance (Eq. \eqref{l2}).

\begin{equation}
\label{cos}
\cos (\textbf{a},\textbf{b}) = \frac{\textbf{a} \textbf{b}}{\|\textbf{a}\| \|\textbf{b}\|} = \frac{ \sum_{i=1}^{n}{\textbf{a}_i\textbf{b}_i} }{ \sqrt{\sum_{i=1}^{n}{(\textbf{a}_i)^2}} \sqrt{\sum_{i=1}^{n}{(\textbf{b}_i)^2}}}
\end{equation}

\begin{equation}
\label{l1}
l1(\textbf{a},\textbf{b} )=\|\textbf{a} -\textbf{b} \|=\sum _{i=1}^{n}|\textbf{a}_{i}-\textbf{b}_{i}|
\end{equation}

\begin{equation}
\label{l2}
l2(\textbf{a},\textbf{b}) =\|\textbf{b} -\textbf{a}\|_{2} = {\sqrt{\sum _{i=1}^{n}(\textbf{b}_{i}-\textbf{a}_{i})^{2}}}
\end{equation}

Furthermore, the back-end authenticates and authorizes the end-users.
This is ensured with the help of JSON Web Tokens (JWT), which are sent in the HTTP header of each request.

\paragraph{User Interfaces}
The website \url{https://logic-mill.net/} features user registration, project presentation, and documentation in a Single Page web Application (SPA). 
It uses the \textit{Vue} \textit{JavaScript} framework\footnote{See \url{https://vuejs.org}.} and consumes the \textit{GraphQL} API provided by the backend. 
\FloatBarrier
\section{Usage\label{Sec::Usage}} 
The general idea behind the user interface is to enable a simple and easy plug-and-play interaction that simplifies the project setup. 
The machinery to use the document encoder model is already available, and the embeddings of large corpora \textit{(Semantic Scholar, EPO, USPTO, WIPO)} are pre-computed and ready to be used. 
Users can do their projects in less time with fewer resources, since there is no need to download the raw data (time, storage), set up the machine learning pipeline (time, CPU/GPU, memory), encode the documents (time), and search through results (time, CPU/GPU, memory, internal storage).

\begin{figure*}
  \caption{Logic Mill Website - Overview\label{fig:lm_overview}}
  \includegraphics[width=\textwidth]{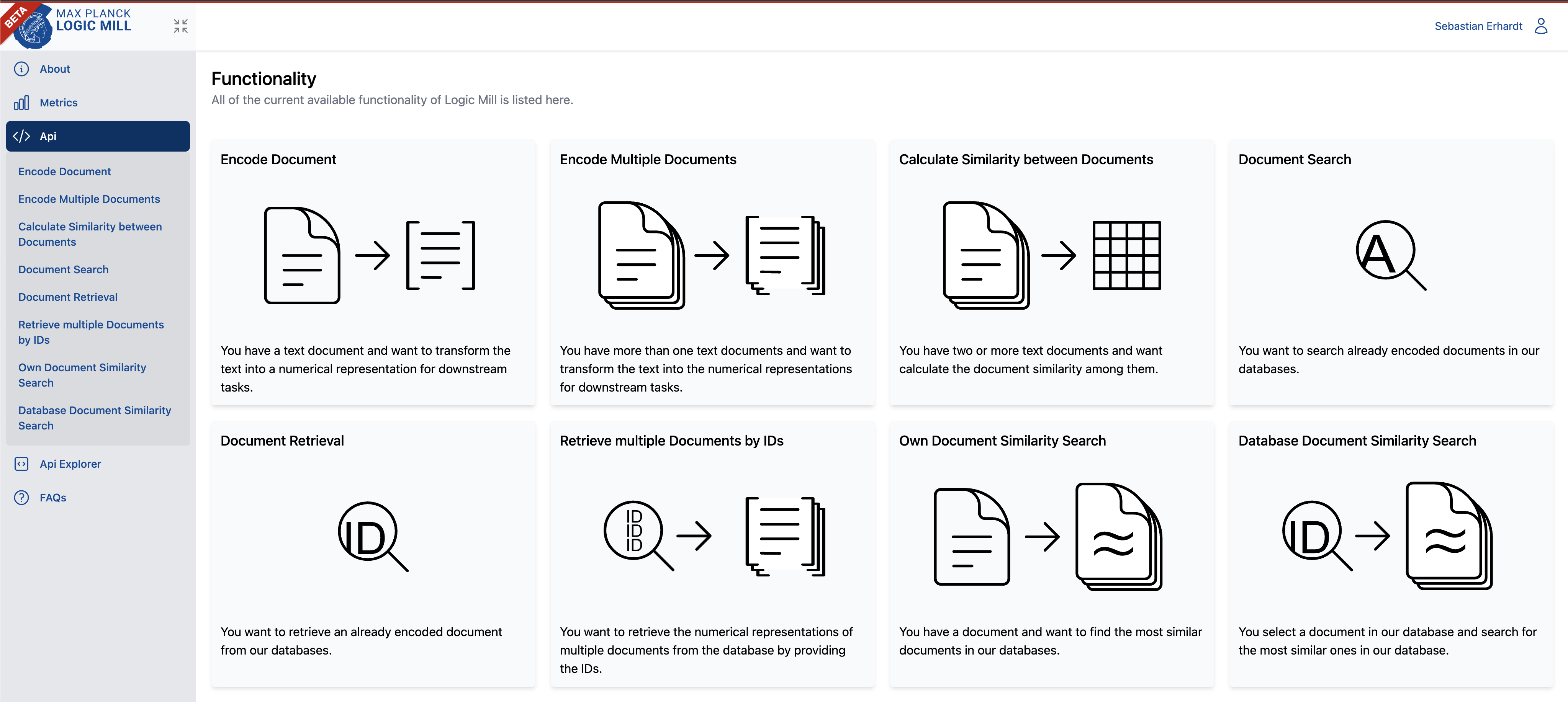}
\end{figure*}

\paragraph{User interfaces}
Upon registration on the website, users can access the system either through the web application on \url{https://logic-mill.net/} (Fig. \ref{fig:lm_overview}) or the Application Programming Interface (API). 

The web application aims to help users familiarize themselves with the queries and data structure. 
It explains the different functionalities of the system and interactively shows the syntactically correct \textit{GraphQL} queries. 
Thus users can experiment, design, and adapt their queries.

The web app auto-generates these queries for various programming languages and tools.
The release version of \textit{Logic Mill} provides these examples in GraphQL for Curl, Python, R, and Go. However, any programming language with the ability to make HTTP requests and retrieve and process JSON responses can interact with the API endpoint.


\paragraph{API Functionality}

\textit{Logic Mill} provides 9 API endpoints for retrieval, pairwise similarity metric computation, and Nearest Neighbor search. Each functionality can be extended to multi-domain corpora (i.e., searching the most similar scientific publications for a patent), and they can involve available documents in the database as well as user-supplied documents or texts. The endpoints are summarized in Table \ref{Table::api_endpoints}.

\begin{table}[ht]
\caption{Overview of \textit{Logic Mill}'s API functionality.\label{Table::api_endpoints}}
  \scalebox{0.86}{\begin{tabular}{l|l|l}
            & Purely database                                                               & with own documents                     \\ \hline
Retrieval   & \begin{tabular}[c]{@{}l@{}}Document, Documents\\ searchDocuments\end{tabular} & encodeDocument, encodeDocuments        \\ \hline
Calculation & similarityCalculation                                                         & encodeDocumentAndSimilarityCalculation \\ \hline
NN Search   & SimilaritySearch                                                              & embedDocumentAndSimilaritySearch      
\end{tabular}
}
\end{table}

\subsection{Retrieval}

\paragraph{Document Retrieval}
In many cases, users can use the system at the beginning of their research projects. 
One can retrieve the pre-computed embeddings for a set of documents, e.g., patent documents or scientific publications, that already exist in the database via the \texttt{Document} and \texttt{Documents} endpoints. 
Users would provide the IDs and corresponding database of the documents along with an indication as to what information they wish to access, e.g. title, abstract, claims, description, authors, inventions, classifications, country, DOI, journal name, and the embedding. 
The information the system will return depends on the document type and the data source. The online API documentation shows the fields that are available in the current version. 

\paragraph{Document Search}
To conduct a keyword-based search, the \texttt{searchDocuments} endpoint can be used. 
\textit{Logic Mill} will retrieve the documents matching the given keywords and metadata from the available corpora. It returns the same information as a retrieval via \texttt{Document} and \texttt{Documents}.

\paragraph{Encode own documents}
The embeddings for own curated documents can be generated and retrieved via the \texttt{encodeDocument} and \texttt{encodeDocuments} endpoints of our API. 
Users provide the title and the abstract, and the document encoder model returns the numerical representation.


\subsection{Calculation}
Users often will be interested in pairwise similarities between documents. 
Although it is possible to retrieve embeddings for a set of documents one-by-one and compute similarity metrics, the system also provides an endpoint for doing precisely the same. 

\paragraph{Calculate Document Similarities}
The endpoint \texttt{similarityCalculation} is used to retrieve the similarity matrix of multiple documents in the database and compare them. 
The input is a list of source documents and target documents (by providing identifiers and indices) as well as the type of distance calculation metric (\texttt{cosine}, \texttt{l1}, \texttt{l2}).

\paragraph{Calculate Similarities with Own Documents}
To retrieve the similarities between a set of own curated documents and documents in the database, users use the \texttt{encodeDocumentAndSimilarityCalculation} endpoint. 
The user provides, for instance, title and abstract of the different documents, identifiers for later reference, and the type of distance calculation metric (\texttt{cosine}, \texttt{l1}, \texttt{l2}). 
The system encodes the documents with the help of the document encoder and uses provided metric to compute the distances among the encoded documents.
The user could then transform the results into a similarity matrix.

\subsection{Nearest Neighbor search}
\paragraph{Database Document Similarity Search}
Given a known document, users can search for the approximate nearest neighbor within the same or other corpora for the most similar documents, based on the \textit{cosine similarity} using the \texttt{SimilaritySearch} query.
For example, the five most similar scientific publications for a specific patent can be requested. 
The title and the similarity score will be returned alongside their IDs within the respective index. 

\paragraph{Own Document Similarity Search}
Users can also provide their own documents and search for similar ones of \textit{Semantic Scholar, EPO, USPTO, WIPO}, that have already been encoded and stored in the database using the \texttt{embedDocumentAndSimilaritySearch} query. 

\section{Conclusion and Future Developments}
\textit{Logic Mill} is a novel software system that helps navigating knowledge embedded in scientific publications, patent documents and other text corpora.
The system is scalable and openly accessible. It is being updated regularly and easy to use. We plan to expand its scope by adding more text corpora such as the English-language Wikipedia and corpus-specific encoders.

Users can leverage the system in the following contexts:
\begin{itemize}
    \item retrieve numerical representations of existing documents in Logic Mill's database
    \item generate numerical representations for their own documents
    \item calculate similarities between users' given documents, or documents in the database, or between users' given document and the one in the database
    \item search for similar documents present in the database given a query document that either exists in the database or users can provide the query document 
\end{itemize}

Researchers interested in innovation, science of science and knowledge transfer may be particularly interested in these capabilities. Its search capabilities may also be of interest to patent examiners and inventors looking for prior art related to their current inventions. Researchers in many fields may want to use \textit{Logic Mill} as a literature and citation recommender system.

\appendix
\label{appendix}

\textit{Logic Mill} provides an API endpoint that uses \textit{GraphQL}.
The \textit{GraphQL} query determines what should be executed and what information should be returned. Users needs an API key to access the API.

\textit{Logic Mill} web app provides a user-interface where dynamic \textit{GraphQL} queries are generated for \textit{cURL, Python, R} and \textit{Go}. \footnote{Stata is another commonly used statistics tool, but cannot retrieve these JSON responses. Recent versions of Stata can, however, embed Python code.}

\begin{figure*}
  \caption{Logic Mill Website - Example Query\label{fig:lm_query_example}}
  \includegraphics[width=\textwidth]{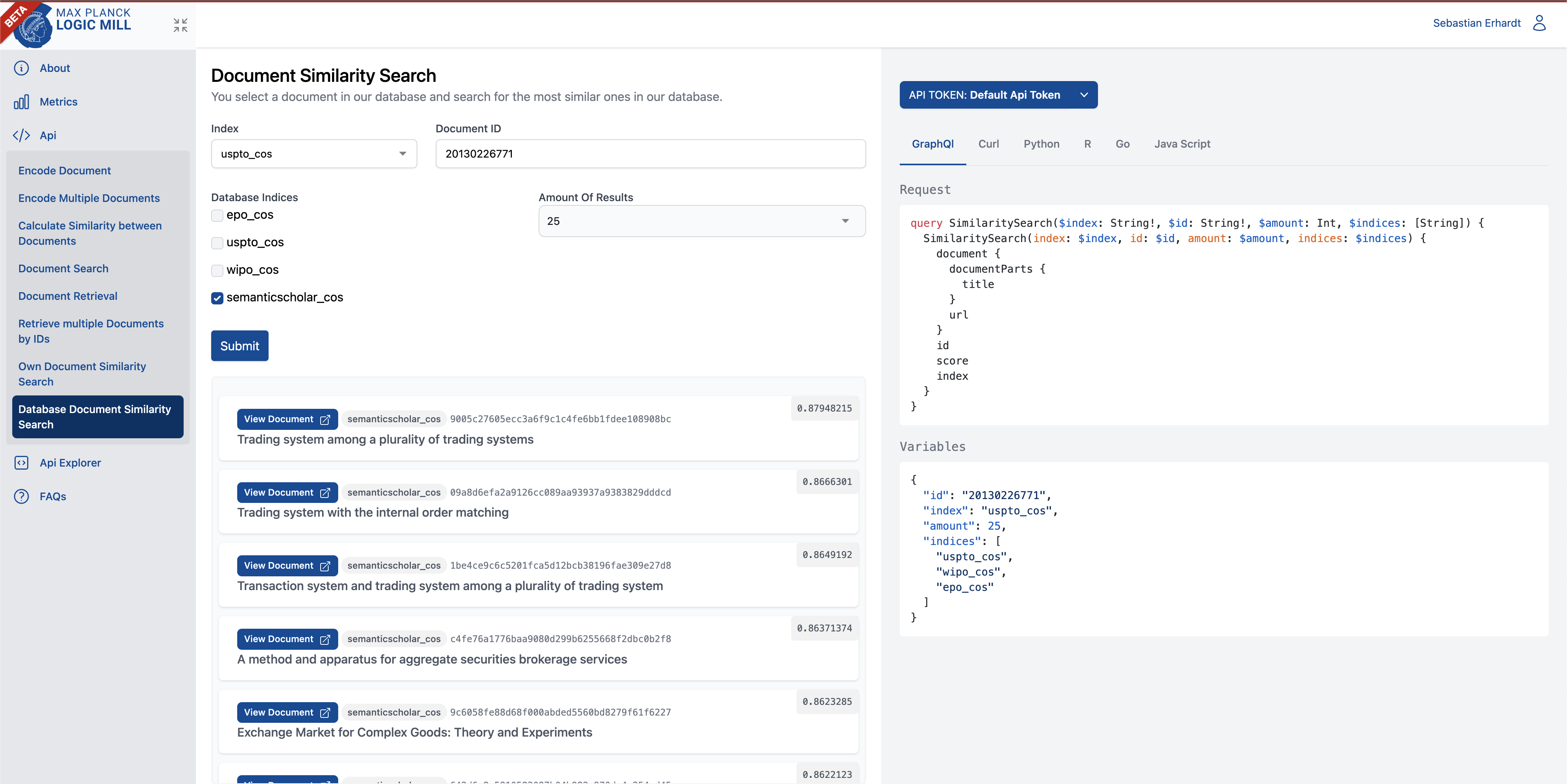}
\end{figure*}

Once a query is constructed in \textit{GraphQL}, it can be implemented and executed using any modern programming language. 

\paragraph{Basic structure of a request}

A basic request written in \textit{Python} is displayed in source code .
\footnote{Our examples are written in Python using the \texttt{requests} package for http requests.} 
The variable called \texttt{query} defines the \textit{GraphQL} query.
User specific \texttt{variables} and requested information.
By providing the index and the ID (in this case an EPO patent with the number of \texttt{EP19164094B1}), the title of the document and the vector (the numerical representation) of the encoded text are requested. 
The return variables can be customized (lines 8-11) to include other information about the document.
The object called \texttt{response} (line 19) is a dictionary that can be processed further.
As one can see, the \textit{Python code} directly includes the \textit{GraphQL} query.

\vfill\break
\paragraph{Example code showing the structure of any interaction with the web API in Python\label{code:basequery}}
\begingroup
    \fontsize{8pt}{10pt}\selectfont
    \begin{Verbatim}[frame=single] 
import requests
import json

TOKEN = 'XXX'
ENDPOINT = 'https://lm/api/endpoint/url/here'
headers = {
  'content-type': 'application/json',
  'Authorization': 'Bearer ' + TOKEN,
}
query="""{
  Document(index: "epo_cos", id: "EP19164094B1") {
    documentParts {
      title
    }
    vector
  }
}"""

r = requests.post(ENDPOINT, 
    headers=headers, 
    json={'query': query})

if r.status_code == 200:
    response = r.json()
    print(response)
else:
    print(f"Error executing\n{query}\non {url}")
    \end{Verbatim}  
\endgroup

\paragraph{Parameters}

While the previous example is the most straightforward and basic implementation, it is not always the most suitable in practice. 
In many cases, one has to make multiple requests to retrieve all the data. 
For example, it is possible to request the encoding for multiple documents in one request, however, it is not possible to retrieve them, for example, 10,000 documents. 
To do this the code needs to include a loop where, with each iteration, a query with different parameters is executed. 
We will call this a parameterized query. 
The user provides two parameters to interact with the web API. 
These are the \textit{GraphQL} query and the query parameters. 
Doing the loop with parameters will make the code more readable. 
The code example \ref{code:parameterized} shows the example with the query and the variables object. 
The code for looping is omitted as well as the base code for handling the request.

\vfill\break
\paragraph{Example code showing with a query with parameters\label{code:parameterized}}
\begingroup
    \fontsize{8pt}{10pt}\selectfont
    \begin{Verbatim}[frame=single] 
# (same setup as above)
# Build GraphQL query
query="""
query Documents($index: String!, $keyword: String!) {
  Documents(index: $index, keyword: $keyword) {
    id
    documentParts {
      title
    }
    vector
  }
}
"""

# Build variables
variables = [
    {"keyword": "EP19164094B1", "index": "epo_cos"},
    {"keyword": "20130226771", "index": "uspto_cos"}
]

# Send request
r = requests.post(ENDPOINT, 
    headers=headers,
    json={
        'query': query,
        'variables': variables
    })

# Handle request
# (...)
    \end{Verbatim}  
\endgroup


\paragraph{Other languages}

Using the \textit{GraphQL} queries in other languages is very similar and the \textit{GraphQL} structure stays the same. 
In Code Sample blow, a query in the R-language is shown. 
The \texttt{ghql} library is used to be able to use the \textit{GraphQL} query directly.

\vfill\break
\paragraph{Example code showing with a query with parameters in R\label{code:r_example}}
\begingroup
    \fontsize{8pt}{10pt}\selectfont
    \begin{Verbatim}[frame=single] 
library(jsonlite)
library(ghql)

URL <-  'https://lm/api/endpoint/url/here' 
variables <-  fromJSON('{
  "data": {
    "id": "ID",
    "parts": [
      {
        "key": "title",
        "value": "Airbags"
      },
      {
        "key": "abstract",
        "value": "Airbags are (...) crash."
      }
    ]
  }
}')

conn <- GraphqlClient$new(
  url = URL,
  headers = list(Authorization = "Bearer <TOKEN>")
)

query <- '
query encodeDocument($data: EncodeObject) {
  encodeDocument(data: $data)
}
'

new <- Query$new()$query('link', query)
res <- conn$exec(new$link, 
variables = variables) %>%
    fromJSON(flatten = F)

res$data
    \end{Verbatim}  
\endgroup


\clearpage
\addcontentsline{toc}{section}{References}
\bibliographystyle{IEEEtran}
\bibliography{./main.bib}

\end{document}